\documentclass[runningheads]{llncs}
\usepackage[T1]{fontenc}
\usepackage{graphicx}
\usepackage{amsmath, amssymb}
\usepackage{booktabs}
\usepackage[table]{xcolor}
\usepackage[hidelinks, colorlinks=true]{hyperref}
\begin{document}
%
\title{SatMap: Revisiting Satellite Maps as Prior for Online HD Map Construction
}
\titlerunning{SatMap}
%
\author{Kanak Mazumder
\orcidID{0009-0006-6806-8388} 
\and
Fabian B. Flohr
\orcidID{0000-0002-1499-3790}
}
\authorrunning{K. Mazumder et al.}
%
\institute{
Munich University of Applied Sciences, Intelligent Vehicles Lab (IVL)\\
Munich, Germany\\
\email{\{kanak.mazumder,fabian.flohr\}@hm.edu} \\
}
\maketitle              
\begin{abstract}
Online high-definition (HD) map construction is an essential part of a safe and robust end-to-end autonomous driving (AD) pipeline. Onboard camera-based approaches suffer from limited depth perception and degraded accuracy due to occlusion. In this work, we propose \textit{SatMap}, an online vectorized HD map estimation method that integrates satellite maps with multi-view camera observations and directly predicts a vectorized HD map for downstream prediction and planning modules. Our method leverages lane-level semantics and texture from satellite imagery captured from a Bird’s Eye View (BEV) perspective as a global prior, effectively mitigating depth ambiguity and occlusion. In our experiments on the nuScenes dataset, \textit{SatMap} achieves $34.8\%$ mAP performance improvement over the camera-only baseline and $8.5\%$ mAP improvement over the camera-LiDAR fusion baseline. Moreover, we evaluate our model in long-range and adverse weather conditions to demonstrate the advantages of using a satellite prior map. Source code will be available \href{https://iv.ee.hm.edu/satmap/}{here}.

\end{abstract}
\section{Introduction}
To ensure safe autonomous driving (AD), an accurate understanding of static road infrastructure, such as lane dividers, boundaries, and crosswalks, is essential. High-definition (HD) map is a high-precision centimeter-level map designed for autonomous driving, containing rich semantic information of road topology and traffic rules~\cite{liao-maptr2023}. In recent years, HD maps have been indispensable for AD development, especially for downstream tasks such as trajectory prediction and planning. However, labeling and maintaining HD maps over time are costly and not scalable, limiting the possibility of autonomous vehicle (AV) deployment~\cite{yuan-streammapnet2024}. In recent years, deep-learning-based approaches for creating online HD maps have gained popularity, which construct HD maps around the ego-vehicle using onboard sensors. 

Early works~\cite{philion-lift2020,li-bevformer2022,liu-bevfusion2022} formulate HD map construction as a semantic segmentation task using the advancement of BEV representation. However, rasterized HD maps lack instance and structure information of map elements and are therefore not compatible for downstream tasks (e.g., motion forecasting and planning). MapTR~\cite{liao-maptr2023} proposes an end-to-end vectorized HD map construction framework using a DETR-based decoder and novel hierarchical query embedding and hierarchical bipartite matching. Recent research in online HD map closely follows MapTR framework and improve the performance using auxiliary losses~\cite{liao-maptrv22024,zhang-gemap2025}, temporal fusion~\cite{yuan-streammapnet2024,chen-maptracker2025,song-memfusionmap2024}, geometric and structural constraints~\cite{zhang-gemap2025,yu-scalablemap2023}, custom attention and decoder~\cite{yuan-streammapnet2024,zhang-gemap2025,yu-scalablemap2023,zhou-himap2024a}, and prior information~\cite{jiang-pmapnet2024,gao-satmap2024,xiong-neural2023,yang-histrackmap2025}. 

\begin{figure}[h]
\includegraphics[width=\textwidth]{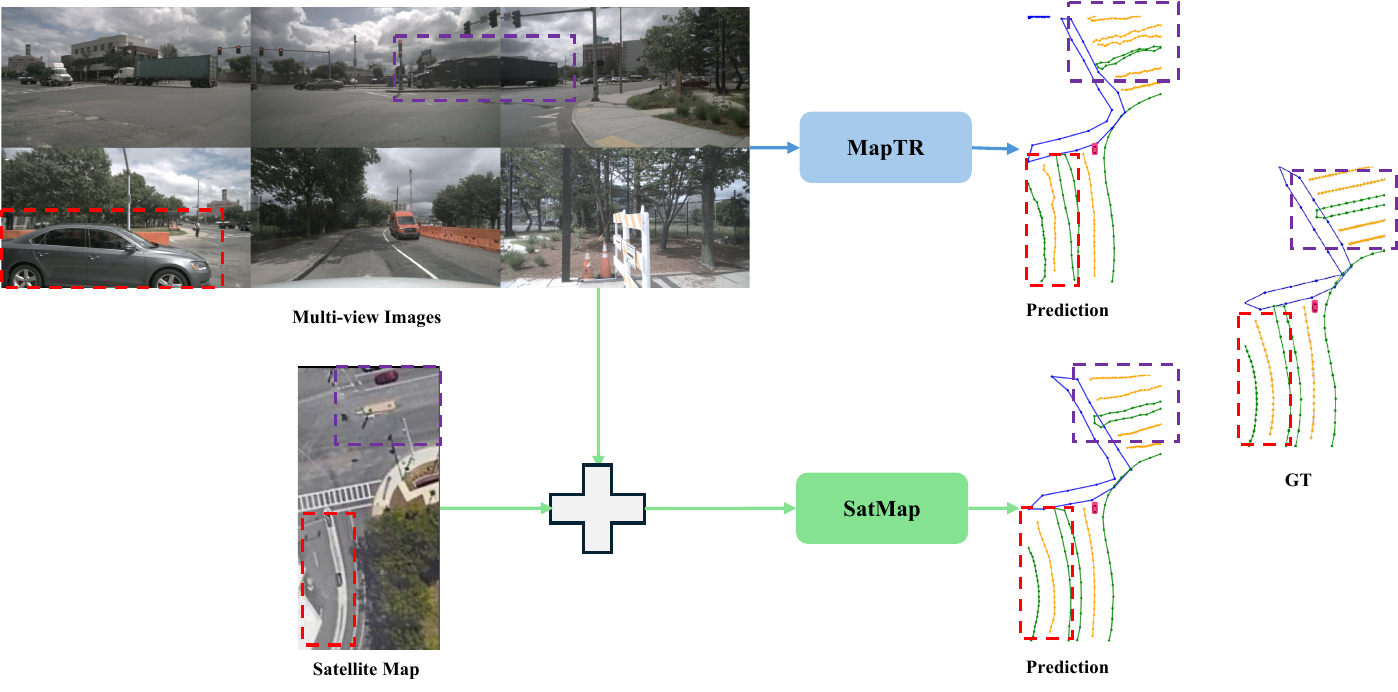}
\caption{We propose SatMap, a camera-satellite fusion-based online vectorized HD map prediction framework. Due to occlusion in perspective view, camera-only models fail to predict accurate map polylines. With a satellite map as prior, SatMap can augment the occluded region with prior information and predict a much accurate HD map. The occluded areas are highlighted in red and magenta.} \label{motivation}
\end{figure}

Most online HD map construction approaches are based on multi-view camera images for cost-effective deployment. Camera-based methods employ a view transformation module to project perspective view (PV) features to BEV space, either using geometric priors~\cite{li-hdmapnet2022,philion-lift2020} or implicit learning~\cite{zhou-crossview2022,li-bevformer2022,liu-petr2022}, due to the lack of depth information. However, the learned BEV representation from PV features is spatially inconsistent and contributes to the noisy map decoding. As the inaccuracy of depth estimation increases with distance, camera-based methods cannot perform well in long-range. Moreover, occlusions from other dynamic agents (e.g., vehicles, pedestrians) and road infrastructure (e.g., parked vehicles, construction) cause inaccurate prediction. Furthermore, challenging lighting (direct sunlight, night) and adverse weather conditions (sunny, rainy, snowy) cause major impediments in camera-based HD mapping. Camera-LiDAR fusion-based methods can alleviate PV-to-BEV transformation inaccuracy and also aid in challenging weather and lighting conditions. StreamMapNet~\cite{yuan-streammapnet2024}, MemFusionMap~\cite{song-memfusionmap2024}, and MapUnveiler~\cite{kim-unveiling} propose to resolve the occlusion issue using effective temporal fusion, which can handle occlusion caused by other moving vehicles to a certain extent, but cannot tackle occlusion caused by static objects (e.g., parked vehicles).

As HD map elements encode static objects in the road infrastructure, which do not change very frequently, it is rational to complement onboard sensor information with prior maps, such as standard-definition (SD) map~\cite{jiang-pmapnet2024,luo-augmenting2024}, satellite map~\cite{gao-satmap2024,ye-smart2025}, HD map from previous traversals~\cite{xiong-neural2023,shi-globalmapnet2024,yang-histrackmap2025}. Given the availability, most prior map fusion-based methods use the SD map to improve performance in long-range. However, SD maps only provide road-level information, as lane width and exact boundary information are missing. Most prior map-based methods only use road skeleton either as a rasterized format~\cite{jiang-pmapnet2024,wu-blosbev2024} or vectorized map embeddings~\cite{luo-augmenting2024,sun-mind2025}. In comparison, a satellite map provides lane-level map information and can tackle the limitations of camera-only methods. 

In this work, we revisit camera-satellite fusion methods for vectorized online HD map prediction and propose SatMap, as shown in Fig.~\ref{motivation}. While previous research~\cite{gao-satmap2024,ye-smart2025} in satellite-based fusion models explores the impact of satellite map as prior, satellite feature extraction and fusion with onboard sensor information are not thoroughly investigated. As a satellite map often contains buildings, trees, and vehicles which occlude boundary and lane information, satellite map features can be noisy and can degrade the quality when fused with onboard BEV features. Moreover, satellite and ego-based observations are only coarsely aligned; this misalignment can further deteriorate the fused feature. ~\cite{gao-satmap2024} adopts a cross-attention mechanism for fusing satellite features extracted using a segmentation network and achieves a minor improvement over the baseline. In contrast, we first adopt a transformer-based backbone for extracting multi-scale satellite map features, which can capture global context and long-range spatial relations even under occlusions. Motivated by the success of BEV level fusion of multi-modal data, we adopt a simple convolution-based fusion module, which can inherently handle local misalignment, alleviating occlusion and small localization errors.  

Our contributions are threefold. First, we propose SatMap, a camera-satellite fusion-based HD mapping framework that efficiently extracts and fuses satellite map features, addressing occlusions and misalignment issues. Secondly, we demonstrate that effectively fusing prior information without major architecture modification can mitigate the limitations of camera-only methods. Finally, we conduct extensive experiments on the nuScenes dataset, demonstrating major performance improvement in long-range and adverse weather conditions.

\section{Related Works}
\subsection{Online Vectorized Map Learning} 
There has been a significant advancement in online vectorized HD map prediction from onboard sensor data. HDMapNet~\cite{li-hdmapnet2022} introduces the first online HD map construction method and uses a heuristic method to post-process pixel-level semantic maps to generate vectorized HD maps. VectorMapNet~\cite{liu-vectormapnet2022} models each map element as a point sequence and predicts them auto-regressively in an end-to-end learning approach. MapTR~\cite{liao-maptr2023} employs permutation-equivalent modeling, and decodes map elements using hierarchical queries and addresses the learning challenges and high latency imposed by the auto-regressive model. StreamMapNet~\cite{yuan-streammapnet2024} addresses temporal inconsistency among single-frame map prediction by adopting a novel streaming feature paradigm. In comparison, MapTracker~\cite{chen-maptracker2025}proposes to address the inconsistency by modeling HD map prediction as a tracking task rather than detection. BeMapNet~\cite{qiao-bemapnet2023a}, PivotNet~\cite{ding-pivotnet2023a}, and GeMap~\cite{zhang-gemap2025} improve the performance of MapTR by adopting effective modeling of map elements and geometric prior ques. ScalableMap adopts structural constraints to explicitly handle challenges of long-range HD map construction. Recent works in HD map learning also explored distillation~\cite{hao-mapdistill2025} and generative~\cite{monninger-mapdiffusion2025} approaches. 

\subsection{Cross-view BEV learning}
The PV-to-BEV view transformation to create a unified BEV representation from camera image features is a significant challenge for 3D perception. In 3D object detection and BEV segmentation domains, PV-to-BEV transformation approaches have been extensively researched. Inverse Perspective Mapping (IPM) is the basic and deterministic method, which uses a homography transformation using known camera parameters. VPN~\cite{pan-vpn2020} learns a common feature representation by transforming image features into BEV space using MLP. LSS~\cite{philion-lift2020} and BEVDet~\cite{huang-bevdet2022} explicitly learns depth distribution for view transformation. CVT~\cite{zhou-crossview2022} utilizes a camera-aware cross-view attention mechanism for mapping individual camera views into the BEV view using geometry-aware positional embedding. GKT~\cite{chen-gkt2022} proposes an efficient and robust view transformation learning approach using a geometry-guided kernel transformer, where camera parameters are used as guidance, making it insensitive to camera deviation. BEVFormer~\cite{li-bevformer2022}, PETR~\cite{liu-petr2022} also adopt learning based approaches for PV-to-BEV transformation using deformable attention and 3D positional embeddings, respectively.

\subsection{Prior-based Map Learning} 
SD maps or Navigation maps can provide a strong prior of road infrastructure and extensively used lane topology reasoning. Given the performance improvement, online map construction methods also adopt them for both rasterized and vectorized representations~\cite{li-local2024}. P-MapNet~\cite{jiang-pmapnet2024} utilizes OpenStreetMap (OSM) road skeleton features and HD map prior features along with onboard sensor data for improved map generation at perception range higher than sensor range. SatforHDMap~\cite{gao-satmap2024} fuses satellite map tiles with onboard camera features hierarchically, consisting of feature-level and BEV-level fusion. SMART~\cite{ye-smart2025} proposes combining the road-level topology priors from SD maps and comprehensive BEV textures from satellite maps and create a unified prior representation. NMP~\cite{xiong-neural2023} stores feature-based global map prior and fuses with onboard sensor data for online map inference and updating the map prior. In comparison, GlobalMapNet~\cite{shi-globalmapnet2024} and HisTrackMap~\cite{yang-histrackmap2025} store predicted maps and utilize historical priors for improving prediction accuracy.

\section{Methodology}
\subsection{Overview}
\begin{figure}
\includegraphics[width=\textwidth]{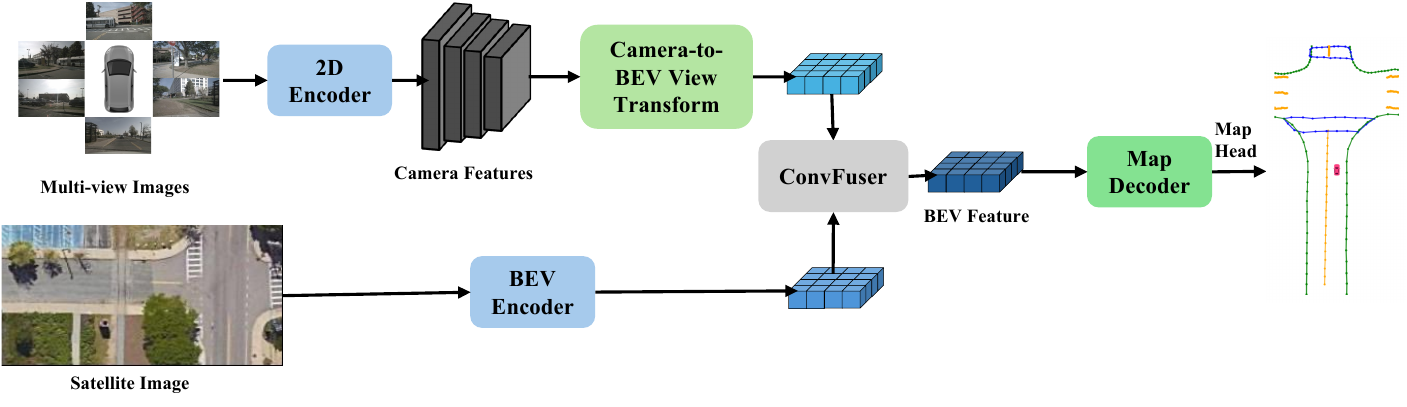}
\caption{The architecture of our proposed SatMap architecture. First, multi-view image features are extracted and projected to BEV space, which are fused with BEV features extracted from the satellite map. The fused BEV feature is used for decoding vectorized map elements.} \label{pipeline}
\end{figure}

Our model utilizes multi-view camera images collected by an autonomous vehicle and the corresponding satellite image extracted using the ego-vehicle pose to create an egocentric HD map. The map is represented as a set of $N$ map instances $\mathcal{M}= \{\boldsymbol{L}_i\}_{i=1}^{N}$, where each map element is represented by a polyline $\boldsymbol{L}_i \in \mathbb{R}^{N_v\times2}$, i.e., an ordered set of two-dimensional points.

As demonstrated in Fig.~\ref{pipeline}, our model architecture comprises four main components: a BEV feature extractor from multi-view camera images, a BEV feature extractor from the cropped satellite image, a multi-modal fusion module, and a DETR-based transformer decoder for predicting map instances from fused BEV features.

\subsection{Multi-modal Feature Extraction}

\subsubsection{Image Feature Extraction}
Given multi-view surround camera images, we aim to construct a unified bird’s-eye-view (BEV) feature representation that preserves geometric consistency across views. Each camera image is first processed by a shared 2D backbone network to obtain multi-scale image features. A set of learnable BEV queries is then initialized on a predefined BEV grid in the ego-centric coordinate system. Using known camera intrinsics and extrinsics, image features are associated with BEV queries through geometry-aware sampling.

Instead of employing multi-view deformable attention as in BEVFormer~\cite{li-bevformer2022}, we adopt GKT~\cite{chen-gkt2022} to perform camera-to-BEV transformation. GKT leverages geometry-guided kernels that explicitly encode camera projection constraints, enabling each BEV query to aggregate image features in a structured and interpretable manner. This design reduces reliance on freely learned offsets and improves geometric consistency across views.

The resulting BEV feature map captures both fine-grained visual appearance and global spatial structure for map element decoding.

\subsubsection{Satellite Map Fetching}
The seminal work~\cite{gao-satmap2024} acquired satellite maps from Google Maps and performed landmark-based coarse alignment for finding the transformation between nuScenes and satellite map coordinates. ~\cite{ye-smart2025} obtained satellite maps covering the nuScenes region from Mapbox Raster Tiles API with a zoom level of 20. OpenSatMap~\cite{zhao-opensatmap2024}, a recent large-scale satellite dataset covering 60 cities, which also claims to include the regions of nuScenes and Argoverse2~\cite{wilson-argoverse22021}, provides high-resolution satellite map tiles with GPS coordinates. We extracted nuScenes samples by converting the ego pose to WGS84 and cropping the ego-centric perception region from merged satellite tiles. In our experiment, we found two scenes from nuScenes, including one validation scene, and a large part of Argoverse2 regions are not included in this dataset. For fair comparison with existing methods, we utilize the complementary nuScenes satellite dataset from~\cite{gao-satmap2024}.

\subsubsection{Satellite Feature Extraction}
We exploit satellite imagery as a source of global, top-down contextual information for BEV map prediction. Owing to its inherent bird’s-eye-view perspective, satellite imagery can be directly encoded into the BEV space without explicit view transformation.

Let $S \in \mathbb{R}^{H_s \times W_s \times 3}$ denote the geo-aligned satellite image centered at the ego vehicle. The satellite image is first processed by a Swin Transformer (SwinT)~\cite{liu-swin2021} backbone to extract multi-scale hierarchical features:
\begin{equation}
\{ \boldsymbol{F}^{\text{sat}}_l \}_{l=1}^{L} = \mathrm{SwinT}(S),
\quad 
\boldsymbol{F}^{\text{sat}}_l \in \mathbb{R}^{H_l \times W_l \times C_l}.
\end{equation}
where l indexes the SwinT stages with progressively decreasing spatial resolution.

To aggregate and enhance multi-scale features, we adopt a Generalized Feature Pyramid Network (GFPN)~\cite{liu-bevfusion2022} as the neck. The GFPN fuses features across different scales through top-down and lateral connections, producing a unified BEV-aligned feature map:
\begin{equation}
\boldsymbol{F}^{\text{sat}}_{\text{bev}} = \mathrm{GFPN}\Big( \{ \boldsymbol{F}^{\text{sat}}_l \}_{l=1}^{L} \Big),
\quad
\boldsymbol{F}^{\text{sat}}_{\text{bev}} \in \mathbb{R}^{H_b \times W_b \times C}.
\end{equation}

The combination of SwinT's shifted-window self-attention and GFPN's multi-scale feature fusion enables effective modeling of both fine-grained road geometry and large-scale topological context. The resulting satellite BEV features provide strong global priors that complement camera-derived BEV representations in downstream vectorized HD map prediction.

\subsubsection{BEV Feature Fusion}
After extracting BEV-aligned features from both cameras and satellite imagery, we fuse them using a simple convolution-based ConvFuser~\cite{liu-bevfusion2022} module. ConvFuser performs convolutional feature integration directly in the BEV space, ensuring spatial alignment between the modalities.  

Specifically, the camera and satellite BEV features are first projected to the same channel dimension using $1\times 1$ convolutions. They are then concatenated along the channel axis and processed through a series of convolutional blocks with residual connections. This enables the network to jointly reason about local appearance cues from cameras and global structural priors from satellite imagery.  

The fused BEV features retain the spatial layout of the BEV grid and provide a rich representation for downstream vectorized map decoding. By fusing the modalities in BEV, ConvFuser avoids complex cross-view transformations while maintaining geometric consistency and improving overall map prediction accuracy.

\subsection{Map Decoder}
After obtaining fused BEV features from camera and satellite modalities, we adopt the DETR-based decoder to predict vectorized map elements in a permutation-invariant manner, as in ~\cite{liao-maptr2023}. Each map element (e.g., lane, road boundary, crosswalk) is represented as an ordered sequence of points:
\[
\mathcal{V} = \{ (c_i, P_i) \}_{i=1}^{N}, \quad P_i = \{ \boldsymbol{p}_{i,j} \}_{j=1}^{M_i},
\]
where $c_i$ is the semantic class of the $i$-th element, and $P_i$ is the ordered set of $M_i$ points in BEV coordinates.

The decoder maintains a set of learnable map queries $\{ \boldsymbol{q}_i^{\text{map}} \}_{i=1}^{N}$, which attend to the fused BEV feature map via multi-head attention. Each query iteratively updates its representation and predicts the corresponding sequence of points through a shared regression head.  

To handle variable-length polylines and unordered set predictions, a set-based Hungarian matching loss is used during training, which matches predicted map elements to ground-truth elements in a permutation-invariant manner. The loss combines classification and point regression components.

The vectorized map decoder thus enables end-to-end, fully vectorized HD map prediction directly from fused BEV features, without relying on rasterized map representations or post-processing steps. Its query-based formulation naturally handles varying numbers of map elements and enforces spatial consistency across the predicted polylines.

\section{Experiments}
\subsection{Dataset}
We conduct our experiments on the nuScenes dataset~\cite{caesar-nuscenes2020}, which provides multi-modal sensor data covering four different cities. It has 1000 scenes of roughly 20 seconds duration each. nuScenes includes six RGB monocular camera images at 12 Hz covering 360 degrees around the ego vehicle, one LiDAR sensor at 20 Hz, and ground-truth HD map annotation at 2 Hz. The HD map contains vectorized polylines of three categories: lane dividers, road boundaries, and pedestrian crossings. The performance of the vectorized map construction task is evaluated using Mean Average Precision (mAP). The Average Precision (AP) is calculated over several Chamfer distances $(\tau \in T, T = \{0.5, 1.0, 1.5\})$  following MapTR~\cite{liao-maptr2023} and then averaged across all thresholds.

\subsection{Implementation Details}
We adopt the MapTR~\cite{liao-maptr2023} training configuration with 24 epochs and a batch size of 32. The model is trained with 4 NVIDIA A40 GPUs. For reducing the computational demand, the dimension of the BEV grid is set to 80$\times$40, i.e., one BEV grid cell corresponds to 0.75m$\times$0.75m perception range. We adopt ResNet50~\cite{he-deep2016} as the camera image backbone and Swin-Tiny~\cite{liu-swin2021} for the satellite map encoder,  both pretrained on ImageNet-1K~\cite{deng-imagenet2009} dataset. We employ GKT~\cite{chen-gkt2022} for the 2D-to-BEV view transformation module.

\begin{table}[h]
\caption{Comparison on the nuScenes \textit{val} split. "C", "L", "SD", and "Sat" denote camera, LiDAR, SD map, and satellite map, respectively. Range refers to the perceived distance along the direction of motion of the ego vehicle. We compare with standard and established long perception ranges and demonstrate the impact of our satellite-based fusion framework. The best result is highlighted in bold.}
\centering
\begin{tabular}{ l c c c c c c c }
\toprule
Method & ~Modality & ~Range & ~Epochs & $\text{AP}_{ped.}$ & $\text{AP}_{div.}$ & $\text{AP}_{bou.}$ & $\text{mAP}$ \\
\hline
MapTR~\cite{liao-maptr2023} & C & [-30.0, 30.0] & 24 & 45.3 & 51.5 & 53.1 & 50.3 \\
MapTR-SDMap~\cite{jiang-pmapnet2024} & C+SD & [-30.0, 30.0] & 24 & 43.7 & 50.9 & 53.5 & 49.4 \\
GeMap~\cite{zhang-gemap2025} & C & [-30.0, 30.0] & 24 & 49.2 & 53.6 & 54.8 & 52.6 \\
SatforHD~\cite{gao-satmap2024} & C+Sat & [-30.0, 30.0] & 24 & 47.2 & 55.3 & 55.3 & 52.6 \\
MapTR-Fusion~\cite{liao-maptr2023} & C+L & [-30.0, 30.0] & 24 & 55.9 & 62.3 & \textbf{69.3} & 62.5 \\
StreamMapNet~\cite{yuan-streammapnet2024} &  C & [-30.0, 30.0] & 24 & - & - & - & 62.9 \\
MGMap~\cite{liu-mgmap2024} & C & [-30.0, 30.0] & 24 & 63.4 & 68.0 & 67.7 & 66.4 \\
HIMap~\cite{zhou-himap2024a} & C & [-30.0, 30.0] & 24 & 62.6 & 68.4 & 69.1 & 66.7 \\
\rowcolor{gray!10}
SatMap(Ours) & C+Sat & [-30.0, 30.0] & 24 & \textbf{67.3} & \textbf{69.1} & 66.9 & \textbf{67.8} \\
\hline
MapTR~\cite{liao-maptr2023} & C & [-30.0, 30.0] & 110 & 56.2 & 59.8 & 60.1 & 58.7 \\
ScalableMap~\cite{yu-scalablemap2023} & C & [-30.0, 30.0] & 110 & 57.3 & 60.9 & 63.8 & 60.6 \\
\rowcolor{gray!10}
SatMap(Ours) & C+Sat & [-30.0, 30.0] & 110  & \textbf{73.9} & \textbf{72.2} & \textbf{72.2} & \textbf{72.7}\\
\hline
MapTR~\cite{liao-maptr2023} & C & [-50.0, 50.0] & 24 & 45.5 & 47.1 & 43.9 & 45.5  \\
\rowcolor{gray!10}
SatMap(Ours) &  C+Sat & [-50.0, 50.0] & 24 & \textbf{63.0} & \textbf{59.7} & \textbf{50.4} & \textbf{57.8} \\
\hline
MapTR~\cite{liao-maptr2023}& C & [-60.0, 60.0] & 110 & 35.6 & 46.0 & 35.7 & 39.1 \\
ScalableMap~\cite{yu-scalablemap2023} & C & [-60.0, 60.0] & 110 & 44.8 & 49.0 & 43.1 & 45.6 \\
\rowcolor{gray!10}
SatMap(Ours) &  C+Sat & [-60.0, 60.0] & 24 & \textbf{61.3} & \textbf{58.3} & \textbf{44.3} & \textbf{54.6} \\
\bottomrule
\end{tabular}
\label{tab:results}
\end{table}

\subsection{Comparison with Baselines}
We compare SatMap with camera-only models and fusion models in the standard 60m$\times$30m perception range. SatMap achieves 34.8\% improvement over the camera-only MapTR~\cite{liao-maptr2023} model and outperforms other advanced vectorized map prediction models such as GeMap~\cite{zhang-gemap2025}, StreamMapNet~\cite{yuan-streammapnet2024}, MGMap~\cite{liu-mgmap2024}, HIMap~\cite{zhou-himap2024a}, as shown in Table~\ref{tab:results}. SatMap performs 8.5\% better than the MapTR-Fusion model. MapTR-SDMap~\cite{jiang-pmapnet2024} and SatforHD~\cite{gao-satmap2024} are MapTR models with SD map and satellite map, respectively. SatMap surpasses MapTR-SDMap and SatforHD by 37.2\% and 28.9\%, respectively. As road boundaries are frequently occluded by trees and buildings, the road features in the fused BEV get blurred, which is responsible for lower improvement in $\text{AP}_{bou.}$. In comparison, LiDAR can provide accurate road geometric information; as a result, MapTR-Fusion achieves the highest $\text{AP}_{bou.}$.

\begin{table}[h]
\caption{We compare our model on the nuScenes dataset under different weather conditions. "sun", "cld", "rny" denote sunny, cloudy, and rainy consecutively.}
\centering
\begin{tabular}{ l | c | c c c c }
\toprule
Method & ~Modality & ~$\text{mAP}_{sun}$ & $\text{mAP}_{cld}$ & $\text{mAP}_{rny}$ & $\text{mAP}_{avg}$ \\
\hline
VectorMapNet~\cite{liu-vectormapnet2022} & C & 43.8 & 44.1 & 36.6 & 41.5 \\
MapTR~\cite{liao-maptr2023} & C & 62.1 & 60.5 & 52.8 & 58.4 \\
BeMapNet~\cite{qiao-bemapnet2023a} & C & 62.5 & 61.9 & 53.0 & 59.1 \\
GeMap~\cite{zhang-gemap2025} & C & 66.0 & 64.3 & 54.4 & 61.5 \\
\rowcolor{gray!10}
SatMap(ours) & C + Sat & \textbf{76.1} & \textbf{80.2} & \textbf{56.2} & \textbf{70.8} \\
\bottomrule
\end{tabular}
\label{tab:weather}
\end{table}

We also compare SatMap with baseline MapTR with a long training schedule and long perception range. SatMap outperforms its baseline in all different settings. For the long-range experiment, we also compare with ScalableMap, as it provides a baseline for long-range HD map prediction. Even with the simplified baseline model architecture, SatMap can outperform specialized architectures for long-range prediction. Due to the lack of open-source implementation of MapTR-SDMap and SatforHD vectorized map, we are not able to compare with them in long-range HD map perception.

\begin{figure}[h]
\includegraphics[width=\textwidth]{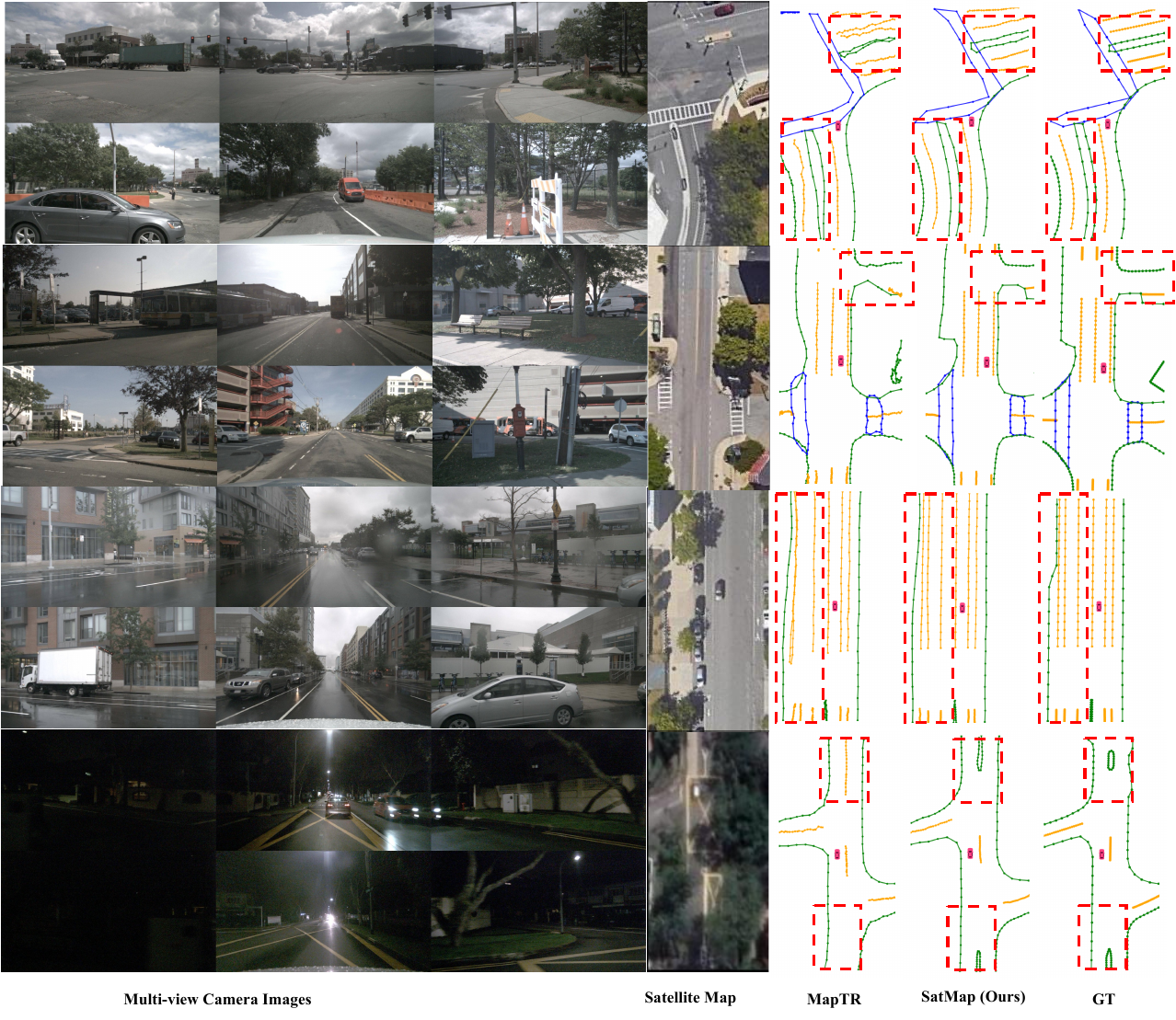}
\caption{Visualization of qualitative results of SatMap in challenging scenes from the nuScenes validation dataset. The left column is the surround views, the middle column is the satellite map, the next columns are inference results of baseline and SatMap, and the right column is the corresponding ground truth. Green lines indicate boundaries, yellow lines indicate lane dividers, and blue lines indicate pedestrian crossings. The challenging regions are indicated in red.} \label{result}
\end{figure}

In order to compare the robustness under different weather conditions, we evaluate SatMap with previous works in Table~\ref{tab:weather}. 
For fair comparison, we evaluate our model trained on 110 epochs under different weather conditions following~\cite{zhang-gemap2025}.
All models show a notable drop in performance in rainy weather. This can be attributed to blur, reflections, water droplets on the camera, and low contrast, all of which obscure visual features. As our model learns to use satellite features as global context to enhance camera features, the degradation of camera features caused by rain cannot be fully compensated. Fig.~\ref{result} shows the qualitative result of SatMap in different weather and light conditions.

\subsection{Ablation Studies}
To examine the efficacy of the proposed SatMap components, we conduct ablation studies on the nuScenes dataset at perception range [-30, 30] meters. Table~\ref{tab:ablation} shows the impact of each component. We use the camera-only MapTR model with 80$\times$40 BEV resolution with ResNet50 backbone and GKT view transformer as our baseline, which achieves 50.1 mAP. We experiment with different satellite map encoders and fusion strategies. First, we adopt ResNet50 as the satellite encoder and utilize vanilla cross-attention between map queries and satellite features for fusion. This fusion technique reaches 53.8 mAP.
With the fused BEV features created using a simple convolution-based fusion module, ConvFuser, the performance reaches 59.1 mAP. Then, we adopt Swin-Tiny as the satellite backbone and experiment with the fusion schemes. With the strong backbone and more refined features, the cross attention-based fusion model improves 57.3mAP (+6.5\%), and the fused BEV-based model attains 67.8 mAP (+14.0\%) improvement over the ResNet50 backbone.

\begin{table}[h]
\caption{Comparison among different satellite feature backbones and fusion strategies. "Swin-T", "Cross-Attn" denotes Swin Transformer Tiny and vanilla cross attention between satellite features and map queries consecutively.}
\centering
\begin{tabular}{ccccc}
\toprule
ResNet50 & ~Swin-T & ~Cross-Attn & ~ConvFuser & mAP \\
\hline
 & & & & 50.1\\
 \checkmark & & \checkmark & & 53.8\\
 \checkmark & & & \checkmark & 59.1\\
 & \checkmark & \checkmark &  & 57.3\\
 \rowcolor{gray!10}
 & \checkmark & & \checkmark & \textbf{67.8}\\
\bottomrule
\end{tabular}   
\label{tab:ablation}
\end{table}

\section{Conclusion}
In this paper, we introduce an effective camera-satellite fusion technique and propose SatMap. SatMap includes robust satellite map feature extraction and fusion with camera features in BEV space, considering occlusion and misalignment between the modalities. SatMap has achieved state-of-the-art performance on the nuScenes dataset over multiple perception ranges and weather conditions. Despite the promising results, SatMap is a single-frame HD map framework, which does not utilize temporal information. Furthermore, with a global lane-level road prior, satellite maps can enable ego-centric online mapping with a monocular camera setting, reducing the computation demand of the mapping module. We expect these findings will inspire further research in this domain. 

\subsubsection{Acknowledgements} This research was conducted within the project “Solutions and Technologies for Automated Driving in Town: An Urban Mobility Project”, funded by the Federal Ministry for Economic Affairs and Climate Action (BMWK), based on a decision by the German Bundestag, grant no. 19A22006N.

%
%

\bibliographystyle{splncs04}
\bibliography{references}

\begin{thebibliography}{10}
\providecommand{\url}[1]{\texttt{#1}}
\providecommand{\urlprefix}{URL }
\providecommand{\doi}[1]{https://doi.org/#1}

\bibitem{caesar-nuscenes2020}
Caesar, H., Bankiti, V., Lang, A.H., Vora, S., Liong, V.E., Xu, Q., Krishnan,
  A., Pan, Y., Baldan, G., Beijbom, O.: nuscenes: A multimodal dataset for
  autonomous driving. In: IEEE/CVF Conference on Computer Vision and Pattern
  Recognition (CVPR). pp. 11618--11628 (2020)

\bibitem{chen-maptracker2025}
Chen, J., Wu, Y., Tan, J., Ma, H., Furukawa, Y.: Maptracker: Tracking with
  strided memory fusion for consistent vector hd mapping. In: IEEE/CVF European
  Conference on Computer Vision (ECCV). vol. 15064, pp. 90--107 (2025)

\bibitem{chen-gkt2022}
Chen, S., Cheng, T., Wang, X., Meng, W., Zhang, Q., Liu, W.: Efficient and
  robust 2d-to-bev representation learning via geometry-guided kernel
  transformer. arXiv preprint arXiv:2206.04584  (2022)

\bibitem{deng-imagenet2009}
Deng, J., Dong, W., Socher, R., Li, L.J., Li, K., Fei-Fei, L.: Imagenet: A
  large-scale hierarchical image database. In: IEEE/CVF Conference on Computer
  Vision and Pattern Recognition (CVPR). pp. 248--255 (2009)

\bibitem{ding-pivotnet2023a}
Ding, W., Qiao, L., Qiu, X., Zhang, C.: Pivotnet: Vectorized pivot learning for
  end-to-end hd map construction. In: IEEE/CVF International Conference on
  Computer Vision (ICCV). pp. 3649--3659 (2023)

\bibitem{gao-satmap2024}
Gao, W., Fu, J., Shen, Y., Jing, H., Chen, S., Zheng, N.: Complementing onboard
  sensors with satellite maps: A new perspective for hd map construction. In:
  IEEE International Conference on Robotics and Automation (ICRA). pp.
  11103--11109 (2024)

\bibitem{hao-mapdistill2025}
Hao, X., Li, R., Zhang, H., Li, D., Yin, R., Jung, S., Park, S.I., Yoo, B.,
  Zhao, H., Zhang, J.: Mapdistill: Boosting efficient camera-based hd map
  construction via camera-lidar fusion model distillation. In: IEEE/CVF
  European Conference on Computer Vision (ECCV). vol. 15061, pp. 166--183
  (2025)

\bibitem{he-deep2016}
He, K., Zhang, X., Ren, S., Sun, J.: Deep residual learning for image
  recognition. In: IEEE/CVF Conference on Computer Vision and Pattern
  Recognition (CVPR). pp. 770--778 (2016)

\bibitem{huang-bevdet2022}
Huang, J., Huang, G., Zhu, Z., Ye, Y., Du, D.: Bevdet: High-performance
  multi-camera 3d object detection in bird-eye-view. arXiv preprint
  arXiv:2112.11790  (2022)

\bibitem{jiang-pmapnet2024}
Jiang, Z., Zhu, Z., Li, P., Gao, H.a., Yuan, T., Shi, Y., Zhao, H., Zhao, H.:
  P-mapnet: Far-seeing map generator enhanced by both sdmap and hdmap priors.
  IEEE Robotics and Automation Letters (RAL) pp. 8539--8546 (2024)

\bibitem{kim-unveiling}
Kim, N., Seong, H., Ji, D., Jang, S.: Unveiling the hidden: Online vectorized
  hd map construction with clip-level token interaction and propagation. In:
  Advances in Neural Information Processing Systems (NeurIPS). vol.~38 (2024)

\bibitem{li-local2024}
Li, J., Jia, P., Chen, J., Liu, J., He, L.: Local map construction methods with
  sd map: A novel survey. arXiv preprint arXiv:2409.02415  (2024)

\bibitem{li-hdmapnet2022}
Li, Q., Wang, Y., Wang, Y., Zhao, H.: Hdmapnet: An online hd map construction
  and evaluation framework. In: IEEE International Conference on Robotics and
  Automation (ICRA). pp. 4628--4634 (2022)

\bibitem{li-bevformer2022}
Li, Z., Wang, W., Li, H., Xie, E., Sima, C., Lu, T., Yu, Q., Dai, J.:
  Bevformer: Learning bird’s-eye-view representation from lidar-camera via
  spatiotemporal transformers. IEEE Transactions on Pattern Analysis and
  Machine Intelligence pp. 2020--2036 (2025)

\bibitem{liao-maptr2023}
Liao, B., Chen, S., Wang, X., Cheng, T., Zhang, Q., Liu, W., Huang, C.: Maptr:
  Structured modeling and learning for online vectorized hd map construction.
  In: International Conference on Learning Representations (ICLR) (2023)

\bibitem{liao-maptrv22024}
Liao, B., Chen, S., Zhang, Y., Jiang, B., Zhang, Q., Liu, W., Huang, C., Wang,
  X.: Maptrv2: An end-to-end framework for online vectorized hd map
  construction. International Journal of Computer Vision (IJCV) pp. 1--23
  (2024)

\bibitem{liu-mgmap2024}
Liu, X., Wang, S., Li, W., Yang, R., Chen, J., Zhu, J.: Mgmap: Mask-guided
  learning for online vectorized hd map construction. In: IEEE/CVF Conference
  on Computer Vision and Pattern Recognition (CVPR). pp. 14812--14821 (2024)

\bibitem{liu-vectormapnet2022}
Liu, Y., Yuan, T., Wang, Y., Wang, Y., Zhao, H.: Vectormapnet: End-to-end
  vectorized hd map learning. In: International Conference on Machine Learning
  (ICML) (2023)

\bibitem{liu-petr2022}
Liu, Y., Wang, T., Zhang, X., Sun, J.: Petr: Position embedding transformation
  for multi-view 3d object detection. In: IEEE/CVF European Conference on
  Computer Vision (ECCV). vol. 13687, pp. 531--548 (2022)

\bibitem{liu-swin2021}
Liu, Z., Lin, Y., Cao, Y., Hu, H., Wei, Y., Zhang, Z., Lin, S., Guo, B.: Swin
  transformer: Hierarchical vision transformer using shifted windows. In:
  IEEE/CVF International Conference on Computer Vision (ICCV). pp. 9992--10002
  (2021)

\bibitem{liu-bevfusion2022}
Liu, Z., Tang, H., Amini, A., Yang, X., Mao, H., Rus, D.L., Han, S.: Bevfusion:
  Multi-task multi-sensor fusion with unified bird's-eye view representation.
  In: IEEE International Conference on Robotics and Automation (ICRA). pp.
  2774--2781 (2023)

\bibitem{luo-augmenting2024}
Luo, K.Z., Weng, X., Wang, Y., Wu, S., Li, J., Weinberger, K.Q., Wang, Y.,
  Pavone, M.: Augmenting lane perception and topology understanding with
  standard definition navigation maps. In: IEEE International Conference on
  Robotics and Automation (ICRA). pp. 4029--4035 (2024)

\bibitem{monninger-mapdiffusion2025}
Monninger, T., Zhang, Z., Mo, Z., Anwar, M.Z., Staab, S., Ding, S.:
  Mapdiffusion: Generative diffusion for vectorized online hd map construction
  and uncertainty estimation in autonomous driving. In: IEEE/RSJ International
  Conference on Intelligent Robots and Systems (IROS). pp. 4099--4106 (2025)

\bibitem{pan-vpn2020}
Pan, B., Sun, J., Leung, H.Y.T., Andonian, A., Zhou, B.: Cross-view semantic
  segmentation for sensing surroundings. IEEE Robotics and Automation Letters
  \textbf{5}(3),  4867--4873 (2020)

\bibitem{philion-lift2020}
Philion, J., Fidler, S.: Lift, splat, shoot: Encoding images from arbitrary
  camera rigs by implicitly unprojecting to 3d. In: IEEE/CVF European
  Conference on Computer Vision (ECCV). vol. 12359, pp. 194--210 (2020)

\bibitem{qiao-bemapnet2023a}
Qiao, L., Ding, W., Qiu, X., Zhang, C.: End-to-end vectorized hd-map
  construction with piecewise b\'ezier curve. In: IEEE/CVF Conference on
  Computer Vision and Pattern Recognition (CVPR). pp. 13218--13228 (2023)

\bibitem{shi-globalmapnet2024}
Shi, A., Cai, Y., Chen, X., Pu, J., Fu, Z., Lu, H.: Globalmapnet: An online
  framework for vectorized global hd map construction. arXiv preprint
  arXiv:2409.10063  (2024)

\bibitem{song-memfusionmap2024}
Song, J., Chen, X., Lu, L., Li, J., Skinner, K.A.: Memfusionmap: Working memory
  fusion for online vectorized hd map construction. arXiv preprint
  arXiv:2409.18737  (2024)

\bibitem{sun-mind2025}
Sun, R., Yang, L., Lingrand, D., Precioso, F.: Mind the map! accounting for
  existing maps when estimating online hdmaps from sensors. In: IEEE/CVF Winter
  Conference on Applications of Computer Vision (WACV). pp. 1671--1681 (2025)

\bibitem{wilson-argoverse22021}
Wilson, B., Qi, W., Agarwal, T., Lambert, J., Singh, J., Khandelwal, S., Pan,
  B., Kumar, R., Hartnett, A., Pontes, J.K., Ramanan, D., Carr, P., Hays, J.:
  Argoverse 2: Next generation datasets for self-driving perception and
  forecasting. In: Advances in Neural Information Processing Systems (NeurIPS).
  vol.~35 (2021)

\bibitem{wu-blosbev2024}
Wu, H., Zhang, Z., Lin, S., Qin, T., Pan, J., Zhao, Q., Xu, C., Yang, M.:
  Blos-bev: Navigation map enhanced lane segmentation network, beyond line of
  sight. In: IEEE Intelligent Vehicles Symposium (IV). pp. 3212--3219 (2024)

\bibitem{xiong-neural2023}
Xiong, X., Liu, Y., Yuan, T., Wang, Y., Wang, Y., Zhao, H.: Neural map prior
  for autonomous driving. In: IEEE/CVF Conference on Computer Vision and
  Pattern Recognition (CVPR). pp. 17535--17544 (2023)

\bibitem{yang-histrackmap2025}
Yang, J., Yang, S., Tan, X., Wang, H.: Histrackmap: Global vectorized
  high-definition map construction via history map tracking. arXiv preprint
  arXiv:2503.07168  (2025)

\bibitem{ye-smart2025}
Ye, J., Paz, D., Zhang, H., Guo, Y., Huang, X., Christensen, H.I., Wang, Y.,
  Ren, L.: Smart: Advancing scalable map priors for driving topology reasoning.
  In: IEEE International Conference on Robotics and Automation (ICRA). pp.
  3298--3304 (2025)

\bibitem{yu-scalablemap2023}
Yu, J., Zhang, Z., Xia, S., Sang, J.: Scalablemap: Scalable map learning for
  online long-range vectorized hd map construction. In: Conference on Robot
  Learning (CoRL). pp. 2429--2443 (2023)

\bibitem{yuan-streammapnet2024}
Yuan, T., Liu, Y., Wang, Y., Wang, Y., Zhao, H.: Streammapnet: Streaming
  mapping network for vectorized online hd map construction. In: IEEE/CVF
  Winter Conference on Applications of Computer Vision (WACV). pp. 7341--7350
  (2024)

\bibitem{zhang-gemap2025}
Zhang, Z., Zhang, Y., Ding, X., Jin, F., Yue, X.: Online vectorized hd map
  construction using geometry. In: IEEE/CVF European Conference on Computer
  Vision (ECCV). vol. 15107, pp. 73--90 (2025)

\bibitem{zhao-opensatmap2024}
Zhao, H., Fan, L., Chen, Y., Wang, H., Yang, Y., Jin, X., Zhang, Y., Meng, G.,
  Zhang, Z.: Opensatmap: A fine-grained high-resolution satellite dataset for
  large-scale map construction. In: Advances in Neural Information Processing
  Systems (NeurIPS). vol.~38 (2024)

\bibitem{zhou-crossview2022}
Zhou, B., Krahenbuhl, P.: Cross-view transformers for real-time map-view
  semantic segmentation. In: IEEE/CVF Conference on Computer Vision and Pattern
  Recognition (CVPR). pp. 13750--13759 (2022)

\bibitem{zhou-himap2024a}
Zhou, Y., Zhang, H., Yu, J., Yang, Y., Jung, S., Park, S.I., Yoo, B.: Himap:
  Hybrid representation learning for end-to-end vectorized hd map construction.
  In: IEEE/CVF Conference on Computer Vision and Pattern Recognition (CVPR).
  pp. 15396--15406 (2024)

\end{thebibliography}

\end{document}